\pdfoutput=1

\documentclass[11pt]{article}

\usepackage[final]{acl}

\usepackage{times}
\usepackage{latexsym}

\usepackage[T1]{fontenc}

\usepackage[utf8]{inputenc}

\usepackage{microtype}

\usepackage{inconsolata}

\usepackage{graphicx}
\usepackage{booktabs}
\usepackage{amsmath}
\usepackage{amssymb}
\usepackage{amsfonts}
\usepackage{array}
\usepackage{pifont}
\usepackage{xcolor}

\DeclareMathOperator*{\argmin}{arg\,min}

\newcolumntype{L}[1]{>{\raggedright\arraybackslash}p{#1}}

\newcommand{\cmark}{\textcolor{green!60!black}{\ding{51}}} 
\newcommand{\xmark}{\textcolor{red}{\ding{55}}}            

\title{Cost-Optimal Grouped-Query Attention for Long-Context Modeling}

\author{Yingfa Chen$^{1*}$, Yutong Wu$^{2*}$, Chenyang Song$^1$, Zhenleng Thai$^1$, \\ \textbf{Xingyu Shen$^1$, Xu Han$^{1\dagger}$, Zhiyuan Liu$^{1\dagger}$, \and Maosong Sun$^1$}  \\
        $^1$NLP Group, DCST, IAI, BNRIST, Tsinghua University, Beijing, China \\ $^2$SIST, University of Science and Technology Beijing, Beijing, China \\ \texttt{chenyingfa1999@gmail.com, wuyutong\_yuna@163.com}
        \\ \texttt{\{han-xu,liuzy\}@tsinghua.edu.cn}
        }

\begin{document}
\maketitle
\begin{abstract}

Grouped-Query Attention (GQA) is a widely adopted strategy for reducing the computational cost of attention layers in large language models (LLMs). However, current GQA configurations are often suboptimal because they overlook how context length influences inference cost. Since inference cost grows with context length, the most cost-efficient GQA configuration should vary accordingly. In this work, we analyze the relationship among context length, model size, GQA configuration, and model loss, and introduce two innovations: (1) we decouple the total head size from the hidden size, enabling more flexible control over attention FLOPs; and (2) we jointly optimize the model size and the GQA configuration to arrive at a better allocation of inference resources between attention layers and other components. Our analysis reveals that commonly used GQA configurations are highly suboptimal for long-context scenarios. Moreover, we propose a recipe for deriving cost-optimal GQA configurations. Our results show that for long-context scenarios, one should use fewer attention heads while scaling up the model size. Configurations selected by our recipe can reduce both memory usage and FLOPs by more than 50\% compared to Llama-3's GQA, with \textit{no degradation in model capabilities}. Our findings offer valuable insights for designing efficient long-context LLMs.\footnote{The code and models are available at \url{https://www.github.com/THUNLP/cost-optimal-gqa}.}

\end{abstract}

\begingroup
\renewcommand{\thefootnote}{\fnsymbol{footnote}}
\footnotetext[1]{Equal contributions.}
\footnotetext[2]{Corresponding authors.}
\endgroup


\section{Introduction}

It is well established that increasing the size of large language models (LLMs) can improve their language modeling qualities \citep{deep-learning-scaling,scaling-law}. Thus, many prior studies have focused on minimizing model size while maintaining quality to ensure cost-effectiveness \citep{chinchilla,minicpm,phi-3}. However, the vast majority of LLMs are Transformer-based \citep{transformer,llama3}, and the cost of running such architectures does not solely depend on the model size. 
Specifically, during inference, a cache of keys/values (i.e., KV cache) is maintained to avoid recomputation in attention layers, resulting in \textbf{memory costs} that scale linearly with the context length. Also, attention layers include the computation of pair-wise attention scores and the weighted summation of value vectors, incurring per-token \textbf{computational costs} that scale linearly with the context length. Many studies have aimed to reduce these costs, including KV cache compression \citep{kv-cache-management-survey}, prompt compression \citep{llmlingua, prompt-compression-survey}, sparse attention \citep{lou2024sparserfastermoreefficient,ge2023model,jiang2024minference}, etc.

One of the most widely used techniques for reducing memory costs is Grouped-Query Attention (GQA) \citep{gqa}, in which attention heads are split into groups and the heads in each group share the same KV vectors. Current implementations of GQA have two critical limitations: (1) Most existing models unnecessarily restrict the total number of head dimensions to be equal to the hidden size, resulting in redundant FLOPs (floating-point operations). (2) When deciding on the number of attention heads and groups, current models do not take into account the influence of context length on the computational and memory costs, resulting in suboptimal long-context configurations.

\begin{figure*}
    \centering
    \includegraphics[width=0.98\linewidth]{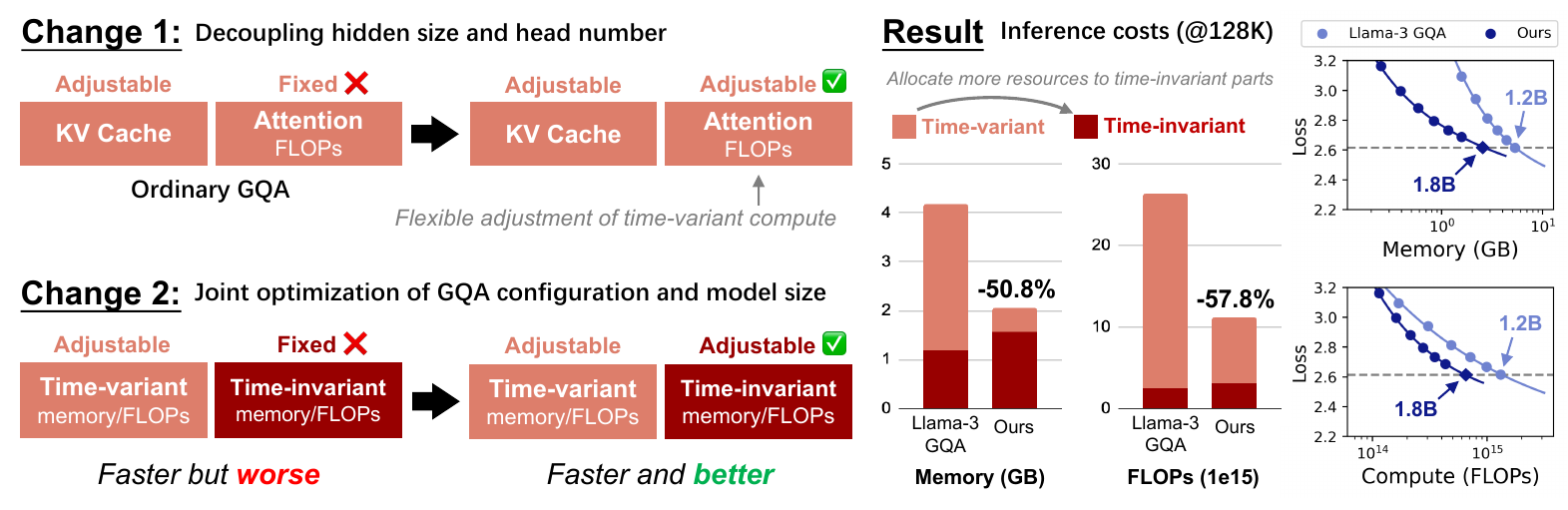}
    \caption{Our approach makes two changes to unlock the flexible adjustment of memory and compute allocation between \textit{time-invariant} components (model weights) and \textit{time-variant} components (KV cache/attention computation). Optimizing resource allocation results in \textbf{cost-optimal GQA configuration} (``Ours''), which has markedly lower memory and FLOPs usage compared to Llama-3, without compromising model capabilities.}
    \label{fig:schema}
\end{figure*}

In this paper, we aim to optimize the cost-effectiveness of GQA Transformers from the perspective of resource allocation. Concretely, we categorize inference costs into \textit{time-invariant costs}, which are constant with respect to context length (e.g., fixed model parameters), and \textit{time-variant costs}, which grow with context length (e.g., attention computation and KV cache).
To freely control the resource allocated to time-variant and time-invariant parts, we make two changes to the existing GQA design procedures: (1) By decoupling the total number of head dimensions and the model hidden size, we unlock a free hyperparameter to control the compute allocated to attention operations. (2) We jointly optimize GQA configurations and model size to modulate the resource allocation between time-variant and time-invariant components. 
After these changes, we can answer our main research question:
\begin{quote}
    \textit{Given an expected inference context length and target loss, how can GQA be configured to minimize inference costs while achieving that loss?}
\end{quote}
To avoid sweeping all combinations of model sizes and GQA configurations, we present a three-step search procedure (detailed in Section~\ref{sec:method}). Our approach is empirically validated on models up to 1.2B parameters. Empirical results show that the widely used Llama-3 GQA configuration \citep{llama3} is highly suboptimal at 128K (which is the context length supported by Llama-3). Instead, our approach gives a configuration that achieves the same loss while reducing inference FLOPs and memory usage by more than 50\%~(Figure~\ref{fig:schema}~(right)).

The contributions of this paper can be summarized by the following points:

\begin{itemize}
    \item By decoupling the model hidden size from the attention head number and jointly optimizing the model size and GQA configuration, we can flexibly allocate memory and compute resources among time-variant and time-invariant components.
    \item We present the first rigorous study to search for the optimal GQA configuration in terms of inference costs for reaching a target loss. Our three-step approach can precisely identify cost-optimal GQA configurations without exhaustively sweeping many configurations.
    \item Our framework reveals valuable insights for designing more cost-effective Transformer LLMs, especially in long-context scenarios.
\end{itemize}

\section{Related Work}
\label{sec:related-work}

\begin{table*}[t]
    \centering
    \footnotesize
    \begin{tabular}{l|l|cc|l}
        \toprule
        & & \multicolumn{2}{c|}{\textbf{Adjustable?}} & \\
        \textbf{Notation} & \textbf{Meaning} & \textbf{Vanilla GQA} & \textbf{This paper} & \textbf{Constrained by} \\
        \midrule
        $T$             & Context length & \xmark & \xmark & \textit{None}\\
        \midrule
        $N$     & Model size  & \xmark & \cmark & \textit{None}\\
        $n_h$   & Attention head number  & \xmark & \cmark & \textit{None} \\
        $n_{kv}$      & KV head number & \cmark & \cmark & \textit{None}\\
        \midrule
        $L$             & Number of layers & \xmark & \xmark & $N$ and pre-defined aspect ratio ($d/L$)\\
        $d$             & Model hidden size & \xmark & \xmark & $N$ and pre-defined aspect ratio ($d/L$)\\
        $d_\text{ff}$   & FFN intermediate size & \xmark & \xmark & $d_\text{ff} \approx 8d/3$\\
        $d_h$           & Head size & \xmark & \xmark & $d_h=64$ \\
        $V$   & Vocabulary size & \xmark & \xmark & Pre-defined vocabulary\\
        \bottomrule
    \end{tabular}
    \caption{Notations in the paper. We optimize more free hyperparameters, resulting in better cost-efficiency.}
    \label{tab:notations-short}
\end{table*}

This paper explores how to build efficient long-context LLMs based on GQA Transformer. Please refer to the LLM-related surveys \citep{zhao2023survey,lu2024small} for more details on LLMs.

\paragraph{Grouped-Query Attention}
The original Transformer model employs multi-head attention (MHA) \citep{transformer}, in which each layer consists of multiple heads that are computed in parallel, and the layer's output is the sum of the heads' outputs. To improve decoding efficiency, especially improving memory efficiency, multi-query attention (MQA) \citep{mqa} shares the weights of all key and value projections among all heads, significantly reducing KV cache size and memory bandwidth requirements during autoregressive decoding. Grouped-query attention (GQA) \citep{gqa} extends this by partitioning heads into groups where each group shares a common KV projection. Formally, MHA is a variant of GQA with independent KV projections per query head, while MQA corresponds to the extreme where all queries share one common KV projection. Recent attention methods based on low-rank factorization, such as MLA \citep{deepseek-v2}, can also be viewed as variants of GQA. Hence, it can be said that most of the current popular LLMs \cite{olmo,pythia,minicpm,llama3,qwen2.5} are built based on GQA. 

\paragraph{Efficient Long-Context Attention}
Attention mechanisms pose a major bottleneck in long-context settings due to high computational and memory costs, especially from the KV cache.
To mitigate this, techniques like sparse attention~\citep{lou2024sparserfastermoreefficient,ge2023model,jiang2024minference}, prompt compression \citep{llmlingua,xiao2023efficient}, and KV cache compression \citep{liu2024kivi,hooper2024kvquant,zhang2024h2o,yao2024sirllm,cai2024lococo} have been proposed. 
While these methods build on and optimize GQA, they often compromise performance relative to vanilla GQA.
Our work focuses on identifying cost-optimal GQA configurations for long-context scenarios through precise characterization of model size, context length, and attention head configurations in terms of their impacts on model performance, computational cost, and memory cost. The efficient long-context attention methods described above remain orthogonal to our GQA architecture search and can be subsequently applied as complementary optimizations to the cost-optimal GQA structures. For more details on efficient long-context attention methods, please refer to the surveys \citep{yuan2024kv,luohe2024keep}.

\paragraph{Scaling Laws for LLMs}
Recent studies on scaling laws for LLMs~\citep{deep-learning-scaling,scaling-law,chinchilla} have established that model loss follows a log-linear relationship concerning model size and training data size. 
They utilize this relationship to minimize the model loss given a fixed training FLOPs budget. However, there are two critical limitations:
(1) These works do not consider the influence of context length on the computational and memory costs.
(2) These laws prioritize the optimal allocation of compute during training, ignoring inference costs. Although \citet{beyond-chinchilla} supplement scaling laws by accounting for total inference FLOPs, their inference cost estimation ignores the influence of context length and memory usage during inference.
Our work extends these studies by accounting for both the computational and memory costs during inference and addressing the impact of context lengths.

\section{Preliminaries: Computational and Memory Costs of GQA Transformers}

In this section, we first briefly introduce GQA Transformers \citep{gqa} and describe key model configurations and their impact on computational and memory costs. Then, we provide a more accurate formula for the computational and memory costs of Transformer-based LLMs that explicitly considers context length and can guide the design of cost-optimal long-context LLMs.
Table~\ref{tab:notations-short} lists the main notations in this paper, and Appendix~\ref{sec:appendix-notations} provides a more complete list.

\subsection{GQA Transformers}
\label{sec:gqa-transformers}

A Transformer model consists of $L$ layers, each of which consists of an attention block and a feedforward network (FFN) block. For each layer, let $\mathbf{x}_i, \mathbf{y}_i \in \mathbb R^{d}$ denote the $i$-th input and output embedding, where $d$ is the model hidden dimension. 

\paragraph{Attention Blocks}
For each head in an attention block, $\mathbf{x}_i$ is first projected into query $\mathbf{q}_i=\mathbf{x}_i\mathbf{W}_q \in \mathbb R^{d_h}$, key $\mathbf{k}_i=\mathbf{x}_i\mathbf{W}_k \in \mathbb R^{d_h}$, value $\mathbf{v}_i=\mathbf{x}_i\mathbf{W}_v \in \mathbb R^{d_h}$, where $d_h$ is the head dimension, then the attention head output is computed as
\begin{equation}
\begin{aligned}
    \mathbf{\tilde h}_{i} = \text{softmax}\left(\frac{\mathbf{q}_i\mathbf{K}_i^\top}{\sqrt{d_h}}\right)\mathbf{V}_i \mathbf{W}_o^\top \in \mathbb R^{d},
    \label{eq:attn}
\end{aligned}
\end{equation}
where $\mathbf{W}_q, \mathbf{W}_k, \mathbf{W}_v, \mathbf{W}_o \in \mathbb R^{d \times d_h}$ are learnable projection matrices. $\mathbf{K}_i^\top=\begin{bmatrix}\mathbf{k}_1^\top \oplus \cdots \oplus \mathbf{k}_i^\top \end{bmatrix}$ and $\mathbf{V}_i^\top=\begin{bmatrix}\mathbf{v}_1^\top \oplus \cdots \oplus \mathbf{v}_i^\top\end{bmatrix}$ are the KV cache for the current attention head, where $\oplus$ denotes the concatenation along the sequence dimension. In MHA Transformers, each attention block consists of $n_h$ heads computed in parallel, and the final attention output $\mathbf h_i\in \mathbb R^d$ is the sum of all head outputs. In GQA, every $n_h/n_{kv}$ query heads share the same KV projection matrices, where $n_{kv}$ is the number of KV heads. 

\paragraph{FFN Blocks}
An FFN block is defined as
\begin{equation}
\begin{aligned}
    \mathbf{y}_i = \sigma \left(\mathbf{h}_{i} \mathbf{W}_\text{up}^\top\right) \mathbf{W}_\text{down} \in \mathbb R^{d},
\end{aligned}
\end{equation}
where $\mathbf{W}_\text{up} \in \mathbb R^{d\times d_\text{ff}}, \mathbf{W}_\text{down} \in \mathbb R^{d_\text{ff}\times d}$ are learnable projection matrices and $\sigma(\cdot)$ is an element-wise activation function.

\begin{table}[!t]
    \small
    \centering
    \begin{tabular}{l|c|c}
        \toprule
        \textbf{Component}  & 
        \textbf{Parameters} & \textbf{Per-token FLOPs} \\
        \midrule
        Input emb.     & $dV$                 & 0 \\
        ATT proj.   & $2Ldd_h(n_h+n_{kv})$ & $4 Ldd_h(n_h + n_{kv})$\\
        ATT comp.   & 0                    & $4 L T  n_hd_h$\\
        FFN          & $2Ldd_\text{ff}$     & $4 Ldd_\text{ff}$ \\
        Output emb.    & 0                    & $2 dV$ \\ 
        \bottomrule
    \end{tabular}
    \caption{Parameters and per-token FLOPs (\textbf{forward pass}) of the main components in Transformers. ``Input emb.'' and ``Output emb.'' represent the input and output embedding layers, respectively, sharing the same embedding weights. ``ATT proj.'' and ``ATT comp.'' represent the projection and computation processes of all attention blocks, respectively.}
    \label{tab:costs-components}
\end{table}

\paragraph{Hyperparameter Constraints}

Let $V$ denote the vocabulary size and $N$ denote the model size. We assume that $d_h$ and $V$ are fixed\footnote{Keeping $d_h$ and $V$ constant for varying model sizes is a common practice. Examples include Llama-3~\citep{llama3} and Qwen3~\citep{qwen3}.}, and $d_\text{ff} \approx 8d/3$, following common LLM design choices~\citep{llama3,olmo,pythia}. For each model size $N$, we assume that the optimal aspect ratio $d/L$ is determined in advance (taken from \citet{pythia}), so each $N$ corresponds to a unique pair $(d, L)$. Table~\ref{tab:notations-short} (right) lists these constraints.

\subsection{Inference Costs of GQA Transformers}
\label{sec:inference-costs}

Table~\ref{tab:costs-components} summarizes the number of parameters for each component in the Transformer model and the FLOPs associated with it. Table~\ref{tab:inference-costs} summarizes the memory and computational costs during inference.

\paragraph{Inference Computational Costs}

$C_{\text{infer}}(T)$ is the number of FLOPs used to process one token within the context with $T$ tokens. This is roughly given as
\begin{equation}
\begin{aligned}
    \label{eq:inference-compute}
    C_{\text{infer}}(T) &= C_\text{const} + C_\text{att}(T) \\
    &= \underbrace{2N}_\text{Time-invariant} + \underbrace{4TL d_h n_h}_\text{Time-variant},
\end{aligned}
\end{equation}
where $C_\text{const}$ denotes the ``time-invariant FLOPs'', the number of FLOPs invariant to the current time step. $C_\text{att}(T)$ denotes the ``time-variant FLOPs'', which is the number of FLOPs used to compute the attention softmax process.

\paragraph{Inference Memory Costs}

$M(T)$ is defined as the memory required to process one token within the context with $T$ tokens. Ignoring the necessary system overhead, we need to store the model parameters and the KV cache, which is roughly:
\begin{equation}
\begin{aligned}
    \label{eq:inference-mem}
    M_{\text{infer}}(T)&=N+N_{kv}(T)\\
    &= \underbrace{N}_\text{Time-invariant} + \underbrace{2TLd_hn_{kv}}_\text{Time-variant},
\end{aligned}
\end{equation}
where $N$ denotes the number of model parameters and $N_{kv}(T)$ denotes the number of values in the KV cache for the context with $T$ tokens.

\begin{table}[t]
    \centering
    \footnotesize
    \begin{tabular}{l|cc}
        \toprule
            \textbf{Type}
            & \textbf{Time-invariant} 
            & \textbf{Time-variant}\\
        \midrule
        FLOPs $(C_\text{infer})$
            & $2N$ 
            & $4TLd_hn_{h}$  \\
        Mem. $(M_\text{infer})$  
            & $N$ 
            & $2TLd_hn_{kv}$ \\
        \bottomrule
    \end{tabular}
    \caption{The time-invariant and time-variant costs of GQA Transformers during inference.}
    \label{tab:inference-costs}
\end{table}

\paragraph{Takeaways}

As listed in Table~\ref{tab:inference-costs}, inference costs can be split into four types: time-invariant FLOPs and memory, and time-variant FLOPs and memory. The time-invariant costs are directly proportional to the model size ($N$), while time-variant FLOPs can be controlled by $n_h$, and time-variant memory can be controlled by $n_{kv}$. Thus, adjusting $N$, $n_h$, and $n_{kv}$ permits fine-grained control over these four kinds of costs. This analysis also implies that a large model may have lower inference costs if its time-variant costs are low enough.

\paragraph{Training Costs}

Since this work mainly focuses on minimizing inference costs, the calculation for the training costs is left to Appendix~\ref{sec:appendix-training-costs}.

\section{Method}
\label{sec:method}

Our objective is to \textit{find the GQA configuration that minimizes inference costs while attaining a given loss}. We approach this by framing the problem as balancing the time-variant and time-invariant costs. In order to unlock the ability to flexibly allocate different amounts of compute and memory to the time-variant and time-invariant components, we make two changes to existing GQA design procedures: (1) We decouple the number of attention heads from the model hidden dimension, and (2) we jointly optimize the model size and the GQA configuration. Figure~\ref{fig:schema} (left) shows the effect of these two changes, and Table~\ref{tab:notations-short} shows the adjustability of different hyperparameters in this work compared to vanilla GQA.

\paragraph{Change 1: Decoupling the Head Number from the Hidden Dimension}

Most existing GQA Transformers adopt $n_h \times d_h=d$, which is arbitrarily chosen in the original Transformer paper \citep{transformer}. This is an unnecessary restriction, rendering GQA unable to adjust the time-variant FLOPs. We decouple $n_h$ from $d$, unlocking a free hyperparameter $n_h$ that controls the number of FLOPs of attention blocks.

\paragraph{Change 2: Joint Optimization of Model Size and GQA Configuration}

In addition to the time-variant costs, we also want to control the time-invariant costs (FFNs, attention QKV/output projections, etc.). 
Specifically, by reducing $N$, but increasing $n_h$, we can allocate more compute to time-variant components. Similarly, we can allocate more compute to time-invariant components by increasing $N$ and decreasing $n_h$. 
This paper aims to identify the optimal allocation of memory and compute between the time-variant and time-invariant components, by jointly tweaking the GQA configuration $(n_h, n_{kv})$ and the model size $N$.

\subsection{Cost-Optimal GQA Search}
\label{sec:cost-optimal-gqa-search}


\paragraph{Objective Formulation}

With the ability to freely adjust the time-variant and time-invariant costs, we formulate the optimization objective as follows,
\begin{equation}
\begin{aligned}
\argmin_{n_h,n_{kv},N} \, & Z(T,N,n_h,n_{kv})\\
\text{ s.t. }& \mathcal L(T, N, n_h,n_{kv}) \le \mathcal L^*\\
\text{where }& Z=\lambda M_\text{infer}^{\alpha}  + (1- \lambda) C_\text{infer}^{\beta},
\end{aligned}
\end{equation}
where $\mathcal L^*$ is the target LM loss, $\mathcal L$ is the model loss, $\lambda\in[0,1],\alpha,\beta \in\mathbb R$ control the trade-off between compute and memory based on deployment constraints\footnote{Although $M_\text{infer}$ and $C_\text{infer}$ have different measurement units, $(\lambda, \alpha, \beta)$ allow us to control the importance of compute and memory resources under a unified metric.}. Setting $\lambda = 1$ minimizes only $M_\text{infer}$, while $\lambda=0$ minimizes only $C_\text{infer}$. We refer to $Z$ as the \textit{hardware-aware cost}. By default, we set $\lambda = 0.9,\alpha=1/2, \beta=1/3$ based on hardware utilization tests in our environment. In other words, the inputs to the optimization objective are $(\mathcal L^*, T)$ and the outputs are $(N,n_h,n_{kv})$.

\paragraph{Influence of Context Length}

We empirically observe that the effect of context length $T$ on loss $\mathcal{L}$ is largely invariant to $N$, $n_h$, and $n_{kv}$ (verified in Section~\ref{sec:influence-of-ctx-len}). This means we can train with moderate context lengths (e.g., $T=8\text{K}$) and extrapolate the loss to longer contexts, saving precious computation resources. However, the influence of model size $N$ and GQA head configuration $H = (n_h, n_{kv})$ on loss is coupled and must be jointly modeled. To this end, we adopt a three-step procedure:

\paragraph{Step 1: Candidate Selection}

Define a candidate set of attention configurations:
\begin{equation}
\begin{aligned}
H_{\text{cand}} &= \{ n_h = 1, 2, 4, \dots, \max(d)/d_h\} \\
& \,\,\,\, \times \{ n_{kv} = 1, 2, 4, \dots, \max(d)/d_h\}\\
& \text{s.t. } n_{kv}\le n_h,
\end{aligned}
\end{equation}
where $\max(d)$ is the hidden size of the largest model used to fit scaling curves in step 2. We round $\max(d)/d_h$ to the nearest power of 2 if necessary.

\paragraph{Step 2: Scaling Curves Fitting}

For each $H \in H_{\text{cand}}$, we train a series of small-scale models with varying $N$ using a sufficiently long context length (we use $T=8\text{K}$), and fit the model loss using a power-law scaling function\footnote{We use the number of non-embedding parameters because it produces more predictable scaling laws in our experiments.} as 
\begin{equation}
\mathcal{L}(N; H) = \left( \frac{a}{N} \right)^b + E,
\end{equation}
where $a$, $b$ are configuration-dependent coefficients and $E$ is the ``natural entropy of language''.

\paragraph{Step 3: Cost Minimization}

For each GQA configuration $H$, we solve for the smallest model size $N^*(H)$ that satisfies the loss constraint as
\begin{equation}
N^*(H) = \frac{a}{\left( \mathcal{L}^* - E \right)^{1/b}}.
\end{equation}
Then, we calculate the inference cost for each configuration and select the one with the lowest cost
\begin{equation}
(N^*(H), H^*) = \argmin_{H} Z(T, N, n_{kv}, n_h).    
\end{equation}

\section{Experiments} 

We first explain the experimental settings (Section~\ref{sec:experimental-settings}).
Then, we present the main results and takeaways (Section~\ref{sec:loss-vs-inference-costs}), followed by the actual cost-optimal GQA configurations derived using our approach (Section~\ref{sec:cost-optimal-gqa-results}) and an analysis of the influence of $n_h$ and $n_{kv}$ on LM loss (Section~\ref{sec:influence-of-query-and-kv-heads}).
After that, we present the results for the setting where total training FLOPs is aligned (Section~\ref{sec:aligning-training-costs}).
Finally, we verify that the effect of $T$ on $\mathcal L$ is largely independent of $N$ and $H$ (Section~\ref{sec:influence-of-ctx-len}). 

\begin{figure*}[!t]
    \centering
    \includegraphics[width=1.0\linewidth]{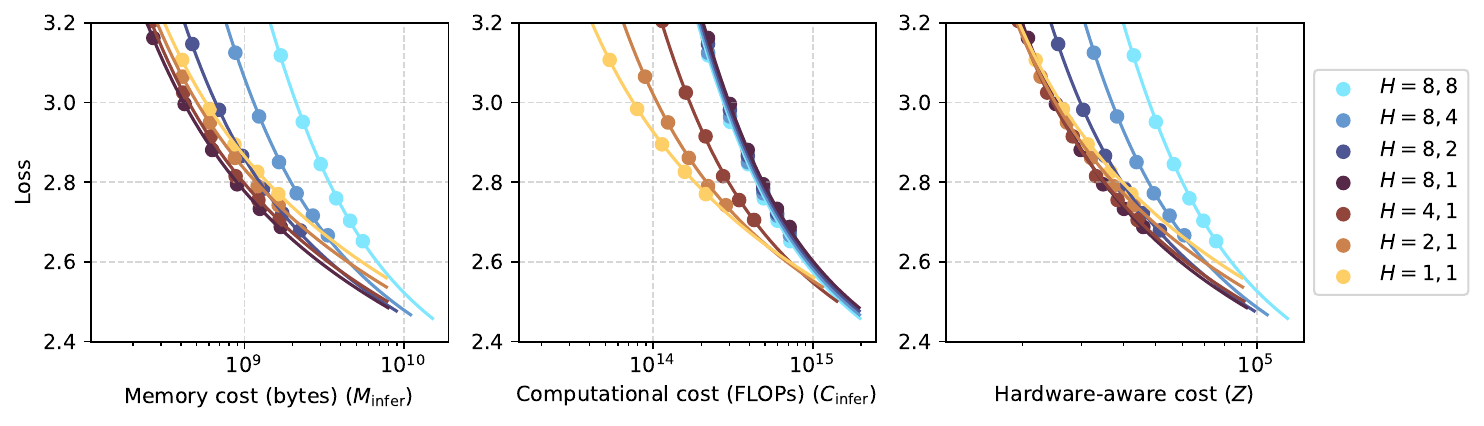}
    \caption{Loss as a function of inference costs with a context length of 128K, assuming we use BF16 for both parameters and the KV cache. $H=(n_h, n_{kv})$ denotes the attention head configuration. $n_h$ and $n_{kv}$ have different effects on the memory cost, computational cost, and loss. $x$-axis is in log scale.}
    \label{fig:loss-vs-costs_diff_heads}
\end{figure*}

\subsection{Experimental Settings}
\label{sec:experimental-settings}

More details of the experimental settings are in Appendix~\ref{sec:appendix-model-config},
Appendix~\ref{sec:appendix-data-processing}, and Appendix~\ref{sec:appendix-training-config}.

\paragraph{Model Configurations}

We adopt the popular Llama-3 \citep{llama3} architecture. For each GQA configuration, we train models from 3M to 1.2B in size. We keep the model configurations as close as possible to \citet{pythia}. We have $\max(d)/d_h=32$, this results in 21 candidate configurations (i.e., $|H_\text{cand}|=21$). 

\paragraph{Data Configurations} 

We use SlimPajama~\citep{slimpj} in our experiments. It is a deduplicated version of the RedPajama corpus~\citep{redpajama} with 627B tokens. In most of our experiments, we use a 20:1 ratio between training data and model parameters, as suggested by \citet{chinchilla}. Additionally, we always ensure that each batch has 512K tokens. 

\paragraph{Training Configurations}

We follow common practices in most of our experiments. We use the AdamW optimizer with the WSD learning rate scheduler \citep{minicpm}. We choose the maximum learning rate by sweeping different values with the MHA model for each model size.

\subsection{Loss vs. Inference Costs}
\label{sec:loss-vs-inference-costs}

Here, we compare the loss-cost tradeoffs of different GQA configurations. 
Figure~\ref{fig:loss-vs-costs_diff_heads} reports the results for a subset of $H_\text{cand}$, showing LM loss as functions of various inference costs ($M_\text{infer}$, $C_\text{infer}$, and $Z$), with a context length of 128K tokens. To save space, we report the result of other context lengths in Appendix~\ref{sec:appendix-influence-of-H-for-different-T}.

\paragraph{Takeaway 1}
We find that loss does not have a simple relationship (e.g., power-plus-constant function) with either memory or computational costs. However, it is possible to predict the loss by fitting the loss as a function of $N$, then transforming the fitted curves along the x-axis to account for the time-variant costs. Fitting loss as a power-plus-constant function of $N$ is highly accurate, with $R^2$ values over 0.999.

\paragraph{Takeaway 2}
The commonly used Llama-3 GQA configuration (i.e., $H=d/d_h, 8$)\footnote{We use ``Llama-3 GQA'' to refer to the GQA configuration on Llama-3 and not the actual publicly released checkpoint, which is trained on huge amounts of proprietary data.} is highly suboptimal at 128K context length. For instance, Llama-3.2-1B uses this head configuration and supports 128K context length. At that length, using $H=(8, 1)$ and increasing the model size to 1.8B would \textbf{achieve the same loss (2.615) while reducing 50.8\% and 57.8\% inference memory and FLOPs usage, respectively} (shown in Figure~\ref{fig:schema} (right)). Alternatively, using $H=8,1$ can achieve a loss that is 0.117 lower than Llama-3.2-1B with the same per-token inference budget in terms of $Z$.


\subsection{Cost-Optimal GQA Configuration}
\label{sec:cost-optimal-gqa-results}

\begin{table}[!t]
    \footnotesize
    \centering
    \begin{tabular}{c|c|c|c|c|c}
        \toprule
        & \multicolumn{5}{c}{\textbf{Expected inference context length ($T$)}}\\
        \midrule
        $\mathcal L^*$ & 8K & 16K & 32K & 64K & 128K \\
        \midrule
        3.0 & $32, 1$ & $16, 1$ & $8,1$  & $4,1$ & $4, 1$ \\
        2.9 & $32, 1$ & $16, 1$ & $16,1$ & $8,1$ & $4, 1$ \\
        2.8 & $32, 2$ & $16, 1$ & $16,1$ & $8,1$ & $8, 1$ \\
        2.7 & $32, 4$ & $16, 2$ & $16,1$ & $16,1$ & $8, 1$ \\
        2.6 & $32, 8$ & $16, 4$ & $16,2$ & $16,2$ & $8, 1$ \\
        2.5 & $32,16$ & $16, 8$ & $16,4$ & $16,2$ & $16, 2$ \\
        2.4 & $32,32$ & $32,32$ & $32,8$ & $32,8$ & $32, 4$ \\
        2.35 & $32,32$ & $32,32$ & $32,32$ & $32,16$ & $32, 8$ \\
        \bottomrule
    \end{tabular}
    \caption{The cost-optimal GQA configuration ($n_h, n_{kv}$) for different target loss $\mathcal L^*$ and context lengths ($T$), while minimizing the \textit{hardware-aware cost} ($Z$, see Section~\ref{sec:cost-optimal-gqa-search}). For reference, the loss of 1B, 3B, and 8B of Llama-3 GQA is 2.615, 2.448, and 2.362, respectively.}
    \label{tab:cost-optimal-gqa}
    \vspace{-1em}
\end{table}

Table~\ref{tab:cost-optimal-gqa} reports the cost-optimal GQA for different expected inference context lengths $T$ and target losses $\mathcal L^*$. When the target loss is high, the model is small, making the time-invariant costs low. Thus, the optimal configuration allocates more resources to the time-invariant part by increasing $N$ and reducing $n_h$ and $n_{kv}$. Similarly, when $T$ is great, the time-variant costs are high, making it more attractive to reduce $n_h$ and $n_{kv}$ more aggressively. The results also indicate that there is nothing especially attractive about the commonly used Llama-3 GQA configuration ($d/d_h, 8$). For certain combinations of $\mathcal L^*$ and $T$, the GQA configuration is cost-optimal. However, for a greater number of combinations, it is sub-optimal. The result implies that the configuration of GQA Transformers should consider the expected inference context length. Directly applying the popular GQA configuration results in wasting hardware resources.

\subsection{Influence of Query and KV Heads}
\label{sec:influence-of-query-and-kv-heads}

\begin{figure}[t]
    \centering
    \includegraphics[width=0.95\linewidth]{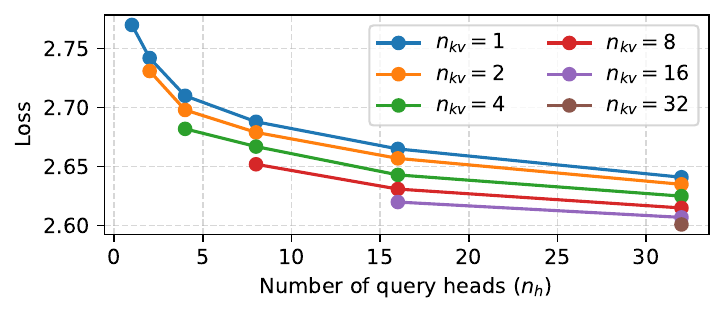}
    \caption{The loss for different number of query heads ($n_h$) and KV heads ($n_{kv}$), with 1.2B model parameters.}
    \label{fig:influence_of_nh_and_nkv}
\end{figure}

Figure~\ref{fig:influence_of_nh_and_nkv} shows the relationship between loss and the number of query heads and KV heads (i.e., different GQA configurations), with a model size of 1.2B. Similar results are observed with other model sizes as well. We emphasize two main takeaways.

\paragraph{Takeaway 1}
The loss reduction by increasing either $n_h$ or $n_{kv}$ exhibits diminishing returns. This means that when $n_h$ or $n_{kv}$ is great, increasing these hyperparameters to reduce loss may not be worth the cost increase. We also found that they exhibit a power-plus-constant relationship (details in Appendix~\ref{sec:appendix-scaling-law-heads}).

\paragraph{Takeaway 2}
Increasing $n_h$ reduces the loss more than increasing $n_{kv}$ by the same amount, although both of them cause the same parameter increase. This means the $n_h$ is more important for model expressivity. Having more query heads allows the model to capture a greater number of dependency patterns. Meanwhile, having more KV heads provides more capacity to store information for each token. The empirical results may indicate that the former is more important for performance.

\subsection{Downstream Performance}
\label{sec:downstream-performance}

\begin{table}[!t]
    \footnotesize
    \centering
    \begin{tabular}{l|cc}
        \toprule
        Evaluation Metric    & $H=32, 8$ & $H=8,1$ \\
                                & (Llama-3 GQA) & (Ours) \\
        \midrule
        Train. throughput (tok/s)       & 18,655 & \textbf{31,260} \\
        Infer. throughput (tok/s)       & 12,921 & \textbf{20,643} \\
        \midrule
        Common-sense  & 45.7\% & 45.5\% \\
        NIAH (1-8K)   & 90.9\% & 96.9\% \\
        NIAH (16K)    & 30.4\% & 46.0\% \\
        NIAH (32K)    & 15.1\% & 18.7\% \\
        NIAH (64K)    &  6.1\% &  7.9\% \\ 
        NIAH (128K)   &  5.2\% &  6.7\% \\
        \bottomrule
    \end{tabular}
    \caption{The throughput of two GQA configurations at 128K context length, and their accuracy on common-sense reasoning (average of 8 tasks) and retrieval tasks (NIAH, varying context length). Although $H=8,1$ has more parameters (1.8B vs. 1.2B), it is much faster for both training and inference.}
    \label{tab:downstream-performance}
\end{table}

Now, we compare the cost-optimal configuration against Llama-3 GQA in terms of training/inference throughput and downstream performance. At $T = $ 128K and $\mathcal L^*=2.615$ (the loss of Llama-3 GQA at 1.2B model size), the cost-optimal GQA configuration is $H=8,1$. Specifically, we train two models, one with $H=32,8$ (Llama-3 GQA) and one with $H=8,1$. Training starts with a 4K context length on 20B tokens. It is then trained with 128K context length for 1B tokens. More training details is given in Appendix~\ref{sec:appendix-downtream-training}.

Training throughput is computed based on the training time while inference throughput is measured with a batch size of 1 on one NVIDIA A800 GPU (with $T=$ 128K). For downstream performance, we evaluate the models on zero-shot common-sense reasoning \citep{eval-harness} and needle-in-a-haystack (NIAH)~\citep{ruler}, which are two widely used LLM benchmarks (more details in Appendix~\ref{sec:appendix-downstream-eval}). The result is shown in Table~\ref{tab:downstream-performance}. One can see that the differences in common-sense reasoning and long-context retrieval are rather small. Meanwhile, the cost-optimal model ($H=8,1$) is much more efficient.

\subsection{Aligning Training Costs}
\label{sec:aligning-training-costs}


\begin{figure}[!t]
    \centering
    \includegraphics[width=0.98\linewidth]{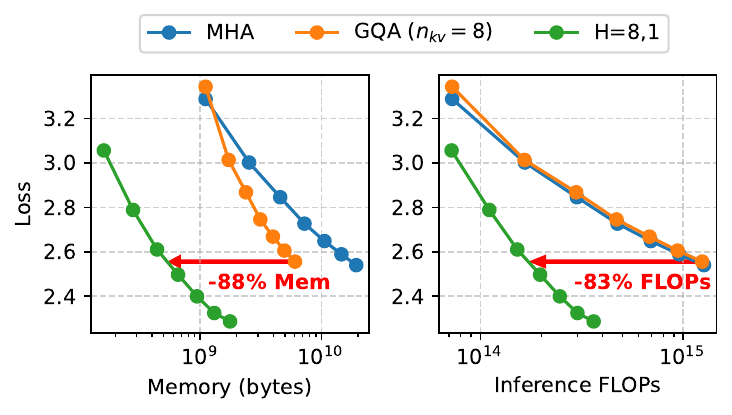}
    \caption{Loss as a function of memory and computational costs, aligned by total training FLOPs at 128K tokens. Each curve is trained with the same amount of training compute.}
    \label{fig:loss_vs_costs_align_training_cost}
\end{figure}

In the previous sections, the training data is always 20 tokens per parameter (i.e., the Chinchilla law). This favors configurations that spend more FLOPs per token. Instead, we can allow more compute-efficient configurations to use more training data to align the training costs of different configurations.

Figure~\ref{fig:loss_vs_costs_align_training_cost} reports the result when we always train with $T=128$K\footnote{LMs are usually trained with short contexts most of the time, so this result may not apply.}. We find that using fewer heads is even more advantageous due to the additional training data, resulting in a model with the same loss but with 88\% and 83\% lower memory and FLOPs usage.

\subsection{Influence of Context Length}
\label{sec:influence-of-ctx-len}

\begin{figure}[!t]
    \centering
    \includegraphics[width=0.98\linewidth]{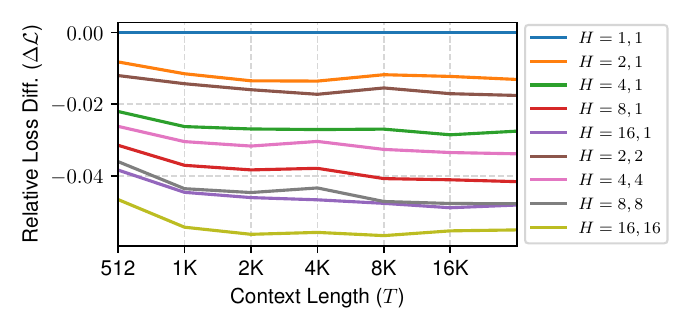}
    \caption{Relative loss difference between various GQA configurations and the $H=1,1$ model, as a function of context length $T$. Model size is 470M.}
    \label{fig:loss_by_T_diff_H}
\end{figure}

\begin{figure}[!t]
    \centering
    \includegraphics[width=0.98\linewidth]{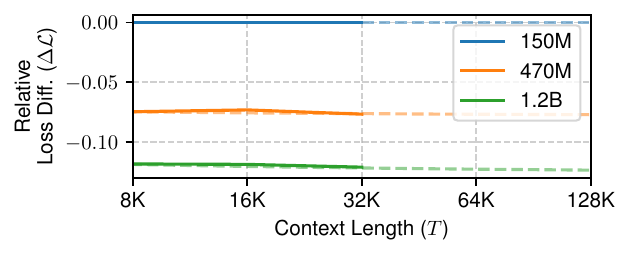}
    \caption{Relative loss difference between varying model size and the 150M model, as a function of context length $T$. These are MHA models.}
    \label{fig:loss_by_T_diff_N}
\end{figure}

In this section, we empirically show that the relationship between context length $T$ and loss $\mathcal L$ is largely invariant to $N$ and $n_h$ when $T$ is sufficiently large. To this end, we measure the relative loss difference between various models and a ``baseline'':
\begin{align*}
\Delta \mathcal L(T) = \frac{\mathcal L(T) - \mathcal L_\text{baseline}(T)}{
\mathcal L_\text{baseline}(T)}
\end{align*}
Figure~\ref{fig:loss_by_T_diff_H} shows the relative loss difference between various GQA configurations with $H=1,1$ as the baseline. Figure~\ref{fig:loss_by_T_diff_N} shows this relationship when varying $N$, with $N$=150M as the baseline. The results show that the relative loss difference is relatively flat when $T>$ 8K (all fluctuations are less than 1\%). The main takeaway is that when applying our cost optimization procedure to longer contexts, we do not have to repeat step 2 (an expensive process) with longer contexts since the loss change of each model will remain roughly the same.

\section{Conclusion}

To optimize the allocation of FLOPs and memory between time-invariant and time-variant components of GQA Transformers, we first decouple the number of attention heads from the model hidden dimensions, enabling a more flexible distribution of FLOPs and memory. Next, we refine the estimation of computational and memory costs in existing approaches by incorporating context length. Our findings reveal that typical configurations of GQA are significantly suboptimal for specific context lengths. Through detailed analysis, we provide valuable insights for improving resource allocation by jointly adjusting the model size and the number of attention heads. As the demand for greater inference context lengths continues to grow, our work marks a critical advancement toward efficient long-context LLMs.

\section*{Limitations}

Like most phenomena in neural language models, we cannot be certain that the conclusions will hold when the models are further scaled up. The power-plus-constant scaling law is also not guaranteed, although it has been empirically validated up to hundreds of billions of parameters. Similarly, there is no guarantee that these laws and our conclusions will hold for an arbitrarily large amount of training data. In general, we have kept our experiments close to research conventions, and the scale of the largest models in our experiments (i.e., 1.2B for Llama-3 GQA and 1.8B for our cost-optimal GQA) is comparable to some real-world LLMs.

We have not thoroughly ablated the influence of all possible hyperparameters due to limited resources. Some important hyperparameters that may affect our conclusions include the head dimension, vocabulary size, and model aspect ratio. More extensive ablation studies are a promising future research direction.

More recent LLMs have adopted even more advanced techniques that complicate the calculations of inference costs. Some notable techniques include speculative decoding, sparse mixture-of-experts, and hybrid recurrence-attention models. Applying our approach to such models remains a promising area for future research.

\section*{Acknowledgments}

This work is supported by the National Key Research and Development Program of China (2024YFB4505603) and a grant from the Guoqiang Institute, Tsinghua University.


\newpage
\bibliography{custom}

\appendix

\newpage
\section{Notations}
\label{sec:appendix-notations}

For completeness, we provide a list of notations we used in the paper, reported in Table \ref{tab:notations}.

\begin{table}[!t]
    \centering
    \footnotesize
    \begin{tabular}{l|p{5.6cm}}
        \toprule
        \textbf{Notation} & \textbf{Meaning} \\
        \bottomrule
        \toprule  
        \multicolumn{2}{c}{\textit{Model hyperparameters}}\\
        \midrule
        $V$ & Vocabulary size, always set to 50,304. \\
        \midrule
        $L$ & Number of layers \\
        \midrule
        $d$ & Model hidden dimension \\
        \midrule
        $d_h$ & Head size, always set to $64$. \\
        \midrule
        $d_\text{ff}$ & FFN intermediate size, we always set $d_\text{ff}=8d/3$. \\
        \midrule
        $\sigma$ & The activation function in FFN \\
        \midrule
        $n_h$ & Number of attention heads \\
        \midrule
        $n_{kv}$ & Number of KV heads (or groups in GQA) \\
        \bottomrule
        \toprule
        \multicolumn{2}{c}{\textit{Inference and Training Costs}}\\
        \midrule
        $C_\text{infer}$ & The computational cost (in FLOPs) per forward pass with a context length of $T$ tokens.\\
        \midrule
        $M_\text{infer}$ & The memory usage (in floating-point values) of serving the model with a context length of $T$ tokens. \\
        \midrule
        $C_\text{train}$ & The computational cost (in FLOPs) used to train the model with a context length of $T$ tokens.\\
        \midrule
        $M_\text{train}$ & The memory usage (in floating-point values) of training the model with a context length of $T$ tokens. \\
        \midrule
        $Z$   & Hardware-aware costs combining both $M_\text{infer}$ and $C_\text{infer}$. Defined in Section~\ref{sec:cost-optimal-gqa-search}.\\
        \bottomrule
        \toprule
        \multicolumn{2}{c}{\textit{Other parameters}}\\
        \midrule
        $T$ & Context length \\
        \midrule
        $N$ & Number of model parameters. \\
        \midrule
        $D_\text{train}$ & Number of training tokens. \\
        \midrule
        $\lambda,\alpha,\beta$ & Hyperparameters controlling the importance of memory and compute resources.\\
        \bottomrule
    \end{tabular}
    \caption{List of notations used in the paper.}
    \label{tab:notations}
\end{table}

\section{Discussions}



\paragraph{What About Other Efficient Attention?}

This paper primarily adjusts the allocation of compute and memory usage by tweaking the model size (controlled with $L$ and $d$) and head configuration $(n_h, n_{kv})$ in GQA, which is a rather simple method. As mentioned, there are many techniques for improving the efficiency of the attention layer, although those have enjoyed less adoption. When using these techniques, the computational and memory costs may be considerably different, and some of our conclusions may not apply. Despite so, our work is still a valuable improvement over existing implementations of GQA.

Recently, Multi-head Latent Attention (MLA) \citep{deepseek-v2} was proposed as a strong alternative to GQA for reducing the KV cache size. During inference, MLA reformulates the attention computation such that all heads share a unified representation for keys and values. In this case, our analysis still applies, since MLA can be seen as a kind of GQA with a different head dimension ($d_h$) and number of attention heads ($n_h, n_{kv}$), and it uses a more complex function to generate the QKV vectors.

\paragraph{What If Context Length Varies?}

The formulas for computational costs (see Table~\ref{tab:costs-full}) are affine functions of $T$, so the \textit{expected costs} are:
\begin{align*}
\mathbb E(C_\text{infer}(T)) &= C_\text{infer}(\mathbb E(T)) \\
\mathbb E(M_\text{infer}(T)) &= M_\text{infer}(\mathbb E(T)) \\
\mathbb E (C_\text{train}(T_\text{train})) &= C_\text{train}(\mathbb E(T_\text{train}))\\
\mathbb E(M_\text{train}(T_\text{train})) &= M_\text{train}(\mathbb E(T_\text{train}))
\end{align*}
where $T_\text{train}$ is the context length during training.
Hence, it suffices to compare the costs with the expected context length. 

\paragraph{Will the Findings Break Down When Scaling Up the Model/Data Size?}

This is a never-ending argument against most neural architectural changes, because no matter the scale of our experiments, we can never be sure that the behavior holds for larger scales. However, our experiments have already covered model sizes up to 1.2B, which is already the size of some widely-used models at the moment \citep{llama3,qwen2.5}. Empirically, it has been widely validated that the scaling law is highly predictable to a good extent beyond the largest model (e.g., Llama-3 accurately predicted the loss of a 405B model with experiments on model sizes up to 16B). Thus, we are confident that our conclusions hold at least for models up to 10B parameters.

\subsection{How to Calculate the Costs of Models of Arbitrary Sizes?} 
\label{sec:appendix-cost-of-arbitrary-size}

\begin{table}[!t]
    \centering
    \begin{tabular}{c|ccc}
        \toprule
        $N$ & $L$ & $d$ \\
        \midrule
        1.2B & 36 & 1536 \\
        1.8B & 36 & 2048 \\
        4B  & 48 & 2560 \\
        6B  & 54 & 3072 \\
        13B & 64 & 4096 \\
        33B & 72 & 6144 \\ 
        64B & 80 & 8192 \\
        \bottomrule
    \end{tabular}
    \caption{The pre-defined configurations used to calculate the aspect ratio of arbitrarily sized models. For models smaller than 1.2B, we use the configurations in Table~\ref{tab:vanilla-model-configs}.}
    \label{tab:aspect-ratio-configs}
\end{table}

In step 3 of our procedure (proposed in Section~\ref{sec:cost-optimal-gqa-search}), we arrive at a critical model size $N^*(H)$. It is a real value, so it does not correspond to an actual model configuration. To calculate the inference costs $(M_\text{infer}, C_\text{infer}, Z)$ of a model of this size, we need $H$ and the aspect ratio of the model $a=d/L$. $H$ is already given, which may be a function of $d$. For the aspect ratio, we perform linear interpolation between the nearest two pre-defined model configurations. The pre-defined model aspect ratios are given in Table~\ref{tab:aspect-ratio-configs}. Then, we use binary search to find the $L$ that corresponds to $N^*(H)$. We can calculate $d$ from $L$ and $a$. Then, we calculate $n_h$ and $n_{kv}$ from $d$ and the specified configuration. With all these values (non-integers) known, we can calculate the model size as well as the inference costs.

To produce an actual model in practice, we suggest simply choosing the configuration $(N,n_h,n_{kv})$ closest to the derived answer in step 3. The slight variations in the performance of the resulting configuration are negligible compared to the huge cost savings gained by selecting the cost-optimal configuration using our approach.

\section{Training Costs of GQA Transformers}
\label{sec:appendix-training-costs}

\begin{table}[t]
    \centering
    \small
    \begin{tabular}{l|cc}
        \toprule
            \textbf{Cost Type}
            & \textbf{Time-invar.} 
            & \textbf{Time-var.}\\
        \midrule
        Infer. FLOPs $(C_\text{infer})$
            & $2N$ 
            & $4TLd_hn_{h}$  \\
        Infer. Mem. $(M_\text{infer})$ 
            & $N$ 
            & $2TLd_hn_{kv}$ \\
        \midrule
        Train. FLOPs $(C_\text{train})$  
            & $6D_\text{train}N$ 
            & $12D_\text{train}TLd_hn_{h}$ \\
        Train. Mem. $(M_\text{train})$
            & $4N$ 
            & $TdL$ \\
        \bottomrule
    \end{tabular}
    \caption{The time-invariant and time-variant costs of GQA Transformers during inference and training.}
    \label{tab:costs-full}
\end{table}

\paragraph{Training Computational Costs}

In addition to inference costs, different head configurations also result in different training costs, because the number of training FLOPs, $C_\text{train}$, is a function of $C_{\text{infer}}$.
Following \citet{scaling-law}, we estimate the FLOPs of the backward pass as double the FLOPs of the forward pass. Let $D_\text{train}$ denote the number of training tokens, $T_i$ denotes the number of tokens preceding the $i$-th training token in the training corpora, then the training FLOPs are:
\begin{align}
C_\text{train} 
    &\approx 3D_\text{train}C_{\text{infer}} \left(\overline T \right) \\
    &= 6D_\text{train} ( 
        N + \underbrace{2L \overline T d_h n_h}_\text{Attention}
    ),
    \label{eq:training-compute}
\end{align}
where $\overline T$ is the average value of $\{T_i|i=1, \cdots, D_\text{train}\}$. When all examples in the training corpora are set to the constant length $T_\text{train}$, during training, we have $\overline T = T_\text{train}/2$. However, in practice, when training long-context LLMs, it is more common to use short contexts for most of the time, and only use long contexts consisting of a small number of tokens to adapt the model to the target context length. Hence, the time-variant FLOPs may only make up a small portion of the training FLOPs, making the cost largely independent of the GQA configuration. Consequently, our paper considers training costs, but focuses more on optimizing inference costs.

\paragraph{Training Memory Costs} 

We only need to store model parameters, activations, gradients, and optimizer states during training. Assuming the widely-used Adam \citep{adam} optimizer without offloading any storage to the CPU, the memory cost is roughly:
\begin{align}
M_\text{train}(T) \approx 4N + \underbrace{TdL}_\text{Activations}.
\end{align}
While it is important to lower the cost of caching activations when $T$ is large, we do not have a free hyperparameter to adjust this cost (like $n_h$ for computational costs and $n_{kv}$ for memory costs). To reduce the size of activations, we have to modify $d$ and/or $L$, which either drastically changes the model size or its aspect ratio. Either of such changes leads to major consequences that are beyond the scope of this paper. Regarding the $4N$ part of training memory cost, it is only dependent on the total model size, so it suffices to minimize the model size, which is already addressed in many existing works \citep{scaling-law,llama3,beyond-chinchilla}.


\begin{figure*}[!t]
    \centering
    \begin{minipage}{0.32\textwidth}
        \centering
        \includegraphics[width=\textwidth]{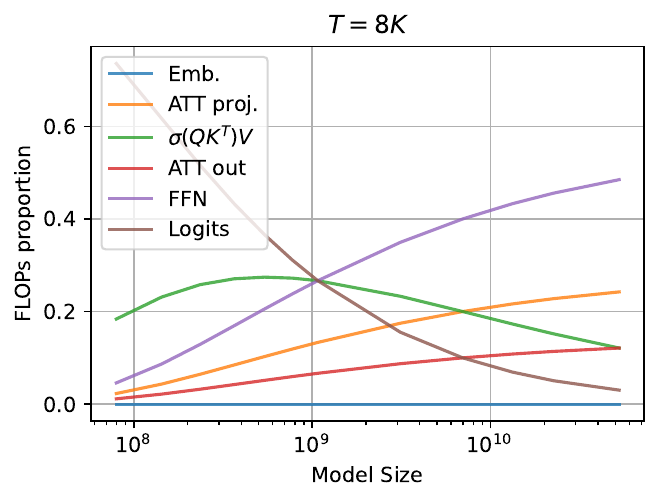}
        \label{fig:flops_distr_subfig1}
    \end{minipage}
    \hfill
    \begin{minipage}{0.32\textwidth}
        \centering
        \includegraphics[width=\textwidth]{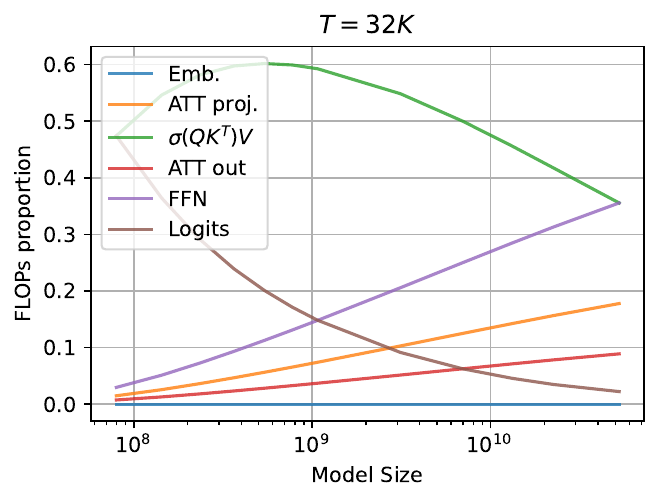}
        \label{fig:flops_distr_subfig2}
    \end{minipage}
    \hfill
    \begin{minipage}{0.32\textwidth}
        \centering
        \includegraphics[width=\textwidth]{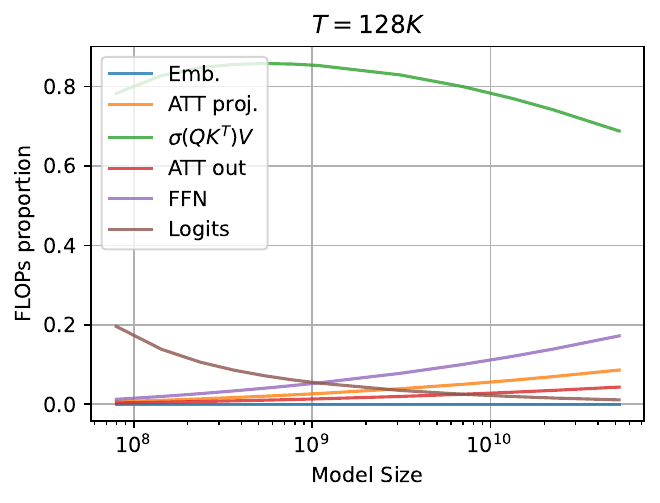}
        \label{fig:flops_distr_subfig3}
    \end{minipage}
    \caption{The proportion of FLOPs allocated to different components in a Transformer LM, with multi-head attention and RoPE. As the context length increases, most FLOPs are spent on the time-variant computation of the attention operator $\sigma(\mathbf Q\mathbf K^\top)\mathbf V$, where $\sigma$ is the row-wise softmax function.}
    \label{fig:flops_distr_by_model_size}
\end{figure*}

\begin{figure*}[!t]
    \centering
    \begin{minipage}{0.32\textwidth}
        \centering
        \includegraphics[width=\textwidth]{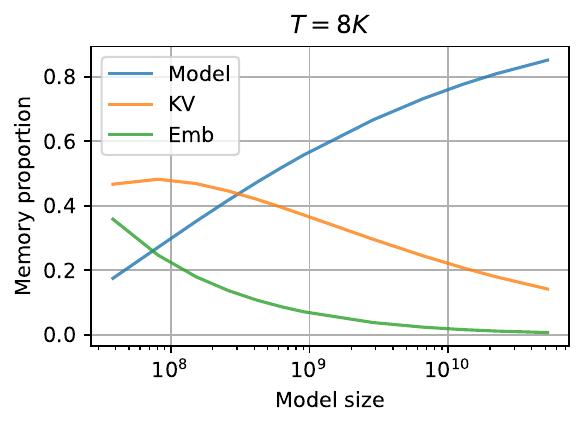}
        \label{fig:mem_subfig1}
    \end{minipage}
    \hfill
    \begin{minipage}{0.32\textwidth}
        \centering
        \includegraphics[width=\textwidth]{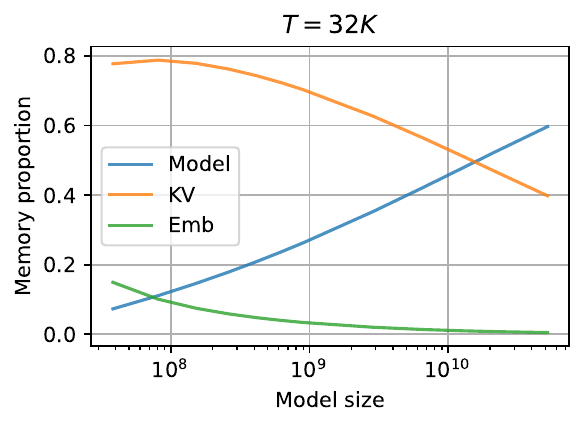}
        \label{fig:mem_subfig2}
    \end{minipage}
    \hfill
    \begin{minipage}{0.32\textwidth}
        \centering
        \includegraphics[width=\textwidth]{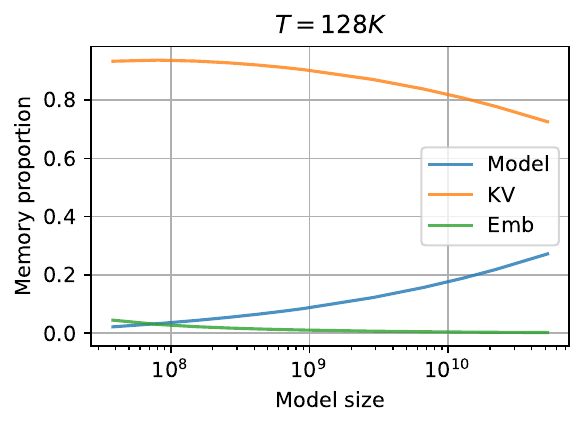}
        \label{fig:mem_subfig3}
    \end{minipage}
    \caption{The proportion of memory allocated to different components in a Transformer LM, with multi-head attention and RoPE. As the context lengths increase, most of the memory usage is spent on storing the KV cache.}
    \label{fig:mem_distr_by_model_size}
\end{figure*}

\section{Model Configurations}
\label{sec:appendix-model-config}

Table~\ref{tab:vanilla-model-configs} shows the configurations of the models in our experiments for fitting the scaling law. In general, we ensure that $d_h=64$, $d_\text{ff} \approx 8d/3$ (rounded to the closest multiple of 32) when scaling the model size, which is adopted from common hyperparameters found in existing LLMs such as GPT \citep{gpt} and Llama \citep{llama3}. We also ensure that the aspect ratio $d/L$ is similar to those used by existing modeling scaling works \citep{pythia,chinchilla,qwen2.5}.
We use the GPT-2 tokenizer, which has a vocabulary size of 50,304, and we tie the input and output embeddings. 

\paragraph{Learning Rate}

The maximum learning rate (LR) is chosen by a grid search on $\{1\times10^i,2\times 10^i,5\times 10^i\mid i=-3, -4, -5\}$ with the vanilla MHA, and choosing the one with best LM loss. Then, we just keep the LR the same across different GQA configurations. While different configurations may have different optimal LR, exhaustively sweeping all LR for each configuration is prohibitively expensive.

\paragraph{Differences From Vanilla GPT}

Compared to the vanilla GPT model \citep{gpt}, we make the following changes to better align with more recent LLMs:
\begin{itemize}
    \item We use RoPE \citep{rope} with a $\theta$ value of 500,000, which is widely used in current LMs \citep{llama3}.
    \item We use SwiGLU FFN instead of the ReLU FFN in GPT.
    \item We use pre-norm \citep{pre-ln} and use RMSNorm \citep{rmsnorm} instead of LayerNorm \citep{layernorm}, which is more common in current LLMs. The epsilon in RMSNorm is $10^{-6}$.
    \item Our model has no bias terms or dropout, which is also common practice and can slightly increase the training efficiency.
\end{itemize}

\begin{table}[!t]
    \centering
    \begin{tabular}{c|ccccc}
        \toprule
        Model size & $L$ & $d$ & $d_h$ & LR \\
        \midrule

        3M   & 4  & 256  & 64 & 1e-3 \\
        19M  & 6  & 512  & 64 & 1e-3 \\
        85M  & 12 & 768  & 64 & 1e-3 \\
        150M & 12 & 1024 & 64 & 1e-3 \\
        200M & 16 & 1024 & 64 & 5e-4 \\
        470M & 24 & 1280 & 64 & 5e-4 \\
        680M & 24 & 1536 & 64 & 2e-4 \\
        1.2B & 36 & 1536 & 64 & 2e-4 \\
        \bottomrule
    \end{tabular}
    \caption{The configurations of the vanilla models with MHA in our experiments, we try to keep it as close to the configurations from \citet{pythia} as possible.}
    \label{tab:vanilla-model-configs}
\end{table}

\section{Data Processing}
\label{sec:appendix-data-processing}

In most of our experiments, we used SlimPajama \citep{slimpj}. We append an EOS token to each document in the corpus before chunking the documents into the specified training length. If the last chunk is shorter than the specified training length, it will be discarded.

\section{Training Configurations}
\label{sec:appendix-training-config}

Here, we provide the default training configurations we used during the experiments.

\begin{itemize}
    \item \textbf{Optimizer:} We use the widely-used AdamW optimizer \citep{adam}, with $\beta_1 =0.9, \beta_2=0.95$, and a weight decay of 0.1. We only apply weight decay to linear layers, which excludes the re-scaling factor in RMSNorm. We also use a gradient clipping value of 1.0.
    \item \textbf{Learning rate scheduler:} We use the warmup-stable-decay (WSD) LR scheduler \citep{minicpm}, with a maximum LR of $5\cdot 10^{-4}$, 10\% warmup steps steps and 20\% decay steps. Warmup starts from 0 and increases linearly to the maximum LR. The decay stage uses a cosine annealing scheme, where the minimum LR is 10\% of the maximum LR.
    \item \textbf{Batch size:} 512K tokens.
    \item \textbf{Floating-point precision:} We use BF16 during training and FP16 during evaluation.
\end{itemize}

\paragraph{Hardware}

All training experiments were run on A800 GPUs, mostly with 8 GPUs.

\section{Memory and Compute Allocations by Model Size}
\label{sec:appendix-memory-and-flops-allocations}

Figure \ref{fig:flops_distr_by_model_size} and \ref{fig:mem_distr_by_model_size} show the FLOPs and memory breakdown of different components as a function of model size. One can see that changes in the model size and/or context length can influence the allocation of FLOPs and memory between different components in the model. For instance, when the context has 128K tokens, the vast majority of FLOPs is spent computing the attention scores and value summation (i.e., $\text{softmax}\left(\mathbf q_i\mathbf K^\top/\sqrt{d_h}\right) \mathbf V$), and the vast majority of memory is spent caching KVs. With 1B model parameters, roughly 90\% of memory will be spent storing the KV cache, and only 10\% will be used to store the model parameters (assuming the KVs and model parameters use the same precision). In other words, the time-variant costs dominate the overall inference costs. Thus, at this context length, we can minimize the overall costs by allocating more resources to the time-invariant components by increasing $N$ and decreasing $n_h$ and $n_{kv}$.

\section{More Results: Loss vs. Inference Costs}
\label{sec:appendix-loss-vs-inference-costs}

\begin{figure*}[!t]
    \centering
    \includegraphics[width=0.95\linewidth]{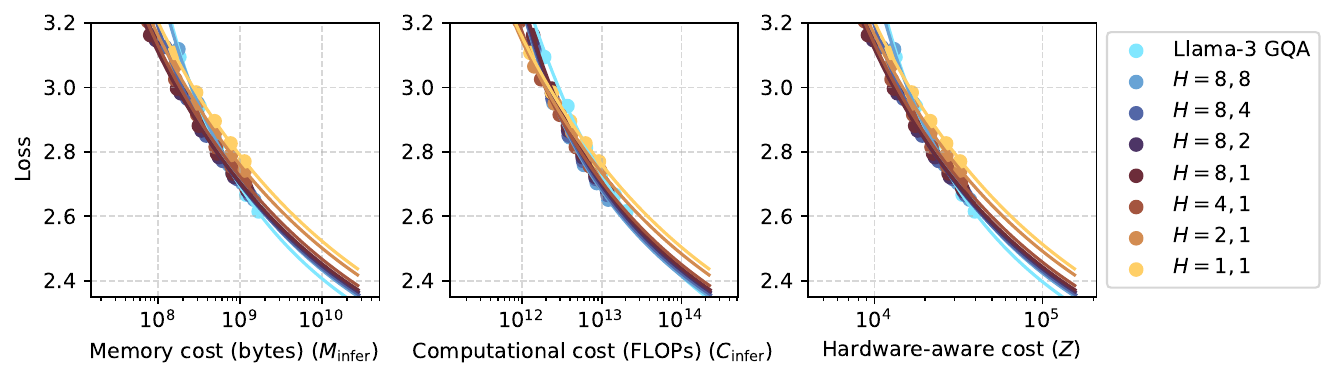}
    \caption{Loss as a function of memory, computational, and hardware-aware ($Z$ in Section \ref{sec:cost-optimal-gqa-search}) costs during inference with a \textbf{context length of 8K tokens}.}
    \label{fig:loss_vs_H_8k}
\end{figure*}

\begin{figure*}[!t]
    \centering
    \includegraphics[width=0.95\linewidth]{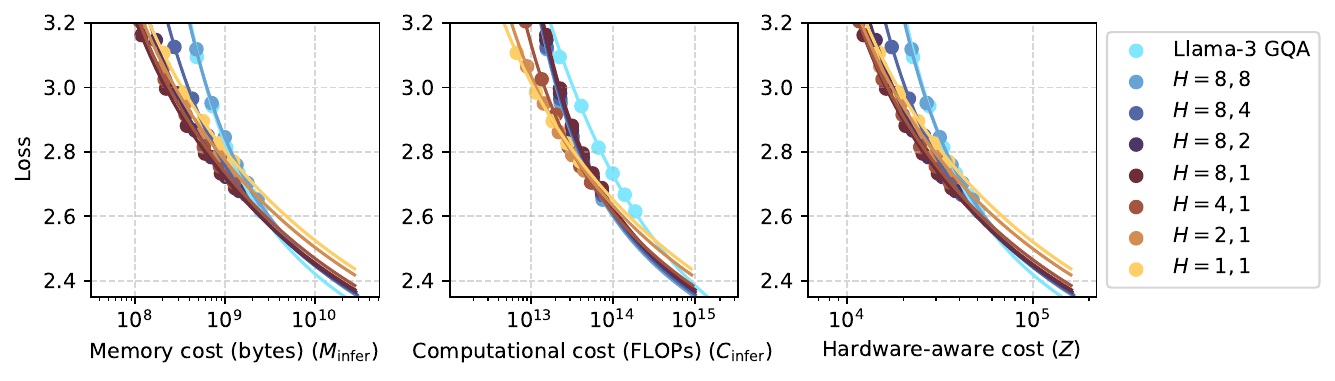}
    \caption{Loss as a function of memory, computational, and hardware-aware ($Z$ in Section \ref{sec:cost-optimal-gqa-search}) costs during inference with a \textbf{context length of 32K tokens}.}
    \label{fig:loss_vs_H_32k}
\end{figure*}

\begin{figure*}[!t]
    \centering
    \includegraphics[width=0.98\linewidth]{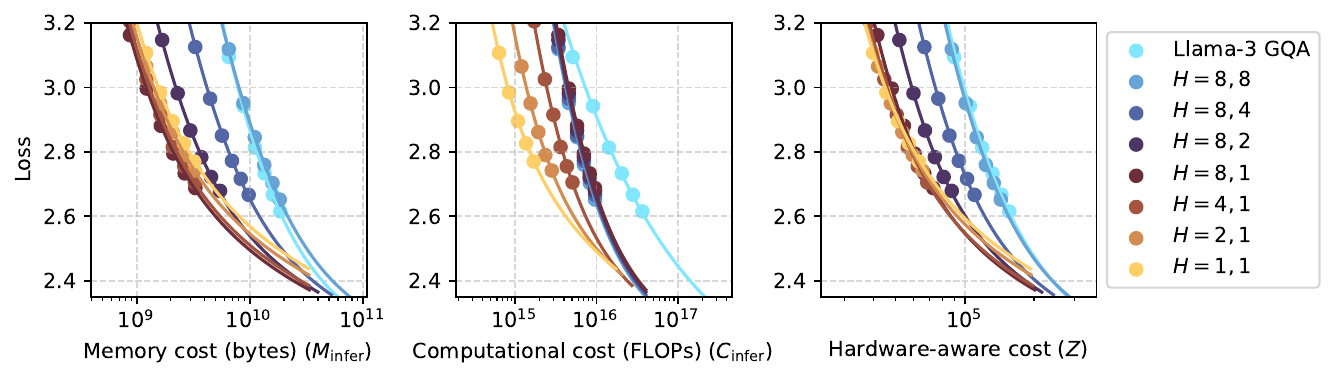}
    \caption{Loss as a function of memory, computational, and hardware-aware ($Z$ in Section \ref{sec:cost-optimal-gqa-search}) costs during inference with a \textbf{context length of 512K tokens}.}
    \label{fig:loss_vs_H_512k}
    \vspace{-1em}
\end{figure*}

Here, we provide the results for the relationship between loss and inference costs for other context lengths. The results are shown in Figure \ref{fig:loss_vs_H_8k}, \ref{fig:loss_vs_H_32k}, and \ref{fig:loss_vs_H_512k}. We can see that for shorter context lengths, the gain of reducing $n_h$ or $n_{kv}$ is relatively small, but the commonly used GQA ($n_{kv}=8$) configuration is still suboptimal at 32K context length. At 1.2B parameters, GQA uses more FLOPs and memory than $H=8,1$. For longer context lengths such as 512K, we can achieve the same loss with less than 10\% of the original memory usage by using fewer KV heads, but a larger model (increasing $N$).

\subsection{Influence of Query and KV Heads for Different Context Lengths}
\label{sec:appendix-influence-of-H-for-different-T}

Here, we provide the supplementary results for Section~\ref{sec:influence-of-query-and-kv-heads} for other context lengths (8K, 32K, and 512K). Similar to the previous section, a greater context length means that the advantage of using fewer heads is greater. In the following section, we explicitly fit the relationship between loss and $n_h$ and $n_{kv}$ with power-plus-constant functions.

\section{The Scaling Laws of Attention Heads}
\label{sec:appendix-scaling-law-heads}

\begin{figure}[t]
    \centering
    \includegraphics[width=1.0\linewidth]{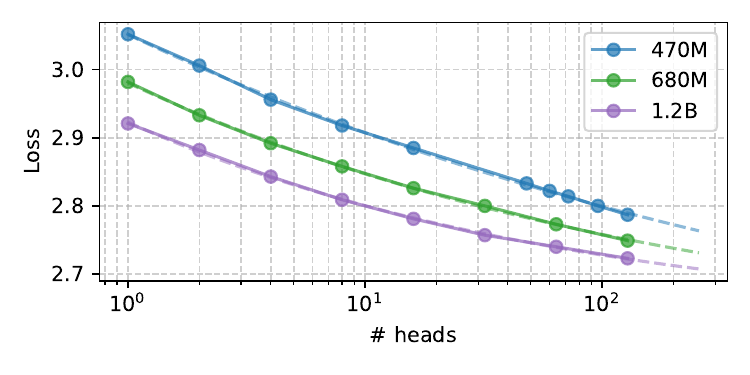}
    \caption{The relationship between LM loss and the number of attention heads, fitted with a power-plus-constant function. The training context length is 1K.}
    \label{fig:loss_vs_heads_diff_N}
\end{figure}

\begin{figure}[t]
    \centering
    \includegraphics[width=1.0\linewidth]{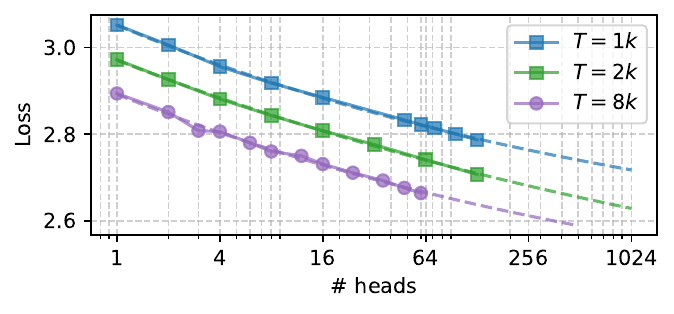}
    \caption{The relationship between LM loss and the number of attention heads, fitted with a power-plus-constant function. The model size is 470M.}
    \label{fig:loss_vs_heads_diff_T}
\end{figure}

In this section, we show that one can predict the loss for a certain head configuration using experiments with a smaller number of heads.
Specifically, we find that---for the first time---the relationship between loss and the number of attention heads closely resembles a power-plus-constant function:
\begin{align*}
    \mathcal L(n_h) = a n_h^b + c
\end{align*}
where $\mathcal L$ is the LM loss, and $a,b,c \in \mathbb R$ are coefficients.
Figure \ref{fig:loss_vs_heads_diff_N} shows that this relationship is observed with different model sizes. The concrete functions for the curves are:
\begin{align*}
    \mathcal L &= 0.579 n_h ^{-0.124} + 2.473 & \text{(470M)} \\
    \mathcal L &= 0.398 n_h ^{-0.177} + 2.583 & \text{(680M)} \\
    \mathcal L &= 0.301 n_h ^{-0.227} + 2.622 & \text{(1.2B)}
\end{align*}
Since the larger model has a greater constant term, this means that these curves will intersect at a certain point (at around $n_h = $ 8K). This is likely incorrect, since the 1.2B model has strictly more parameters than the other models (although at such large values of $n_h$, the relative difference in model size is very small). This means that the fitted curves will break down before $n_h =$ 8K. Fortunately, virtually all LLMs with open weights have fewer than 128 heads, and the fitted curves are very accurate up to 128 heads with $R^2$ values over 0.999. Thus, we conclude that the law is empirically accurate for the vast majority of openly available LLMs.

Similarly, Figure~\ref{fig:loss_vs_heads_diff_T} shows that this trend is consistent across different context lengths. The fitted curves are
\begin{align*}
    \mathcal L &= 1.513 n_h^{-0.039} + 1.53 & (T=1K) \\
    \mathcal L &= 1.436 n_h^{-0.041} + 1.53 & (T=2K) \\
    \mathcal L &= 1.356 n_h^{-0.044} + 1.53 & (T=8K)
\end{align*}
When $n_h$ approaches infinity, the model parameters will be dominated by the attention projection matrices (i.e., QKVO projections). Hence, they converge to the same constant term, which is known as the ``natural entropy of language''. During curve fitting, this constant term is chosen to minimize to fitting error, and we arrive at 1.53. The $R^2$ values of these fits are over 0.999.

From these results, we conclude that this power-plus-constant scaling law between loss and the number of heads is exhibited independently of model size and context length. One important implication of this result is that increasing the number of heads to improve model quality gives diminishing returns. This means that beyond a certain point, the loss reduction brought by further increasing the number of heads is not worth the cost increase.


\subsection{Constant Number of KV Heads}

\begin{figure}[t]
    \centering
    \includegraphics[width=0.98\linewidth]{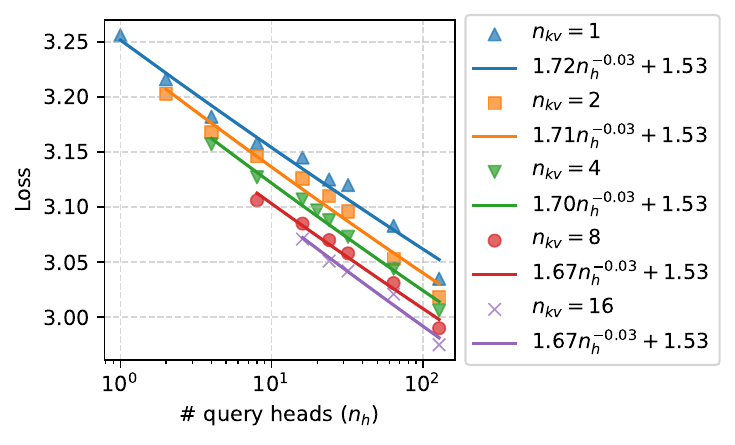}
    \caption{The relationship between loss and $n_h$ when $n_{kv}$ is constant. Model size is 150M.}
    \label{fig:loss_vs_nh_constant_nkv}
\end{figure}

Some LMs (e.g., Llama-3 \citep{llama3}) keep the number of KV heads constant when scaling up the model. Therefore, we also investigate the relationship between LM loss and $n_h$ when $n_{kv}$ is constant. Figure \ref{fig:loss_vs_nh_constant_nkv} shows this relationship with different values of $n_{kv}$ and two model sizes. 
We discover that the relationship is still a power-plus-constant law, but the fitted curves are notably less accurate, with $R^2$ values over 0.97. It is worth noting that the increase in fitting error compared to Section \ref{sec:appendix-scaling-law-heads}) may be attributed to the use of a smaller model (150M vs. 470M).

\section{Experimental Details: Downstream Performance}

This section provides details for Section~\ref{sec:downstream-performance}.

\subsection{Training}
\label{sec:appendix-downtream-training}

The training run for both the Llama-3 GQA and $H=8,1$ (cost-optimal GQA) models are exactly the same. It consists of two phases. The first phase uses the same settings as the scaling experiments in Section~\ref{sec:experimental-settings}. After 20B tokens, we continue training with 128K context length for 1B tokens, using new optimer states. This phase uses a lower maximum LR of 1e-5 for stability and to avoid catastrophic forgetting.

\subsection{Evaluation}
\label{sec:appendix-downstream-eval}

Here, we provide more details regarding the downstream task performance evaluation in Section~\ref{sec:downstream-performance}. We use LM-Evaluation-harness~\citep{eval-harness} for common-sense reasoning, and the needle-in-a-haystack tasks from RULER~\citep{ruler}. For both of these tasks, we evaluate the last four checkpoints of the model, and report the average score of it. This is for reducing the randomness in the results.

\paragraph{Common-Sense Reasoning Tasks} 

We use the popular LM-Evaluation-Harness~\citep{eval-harness} for evaluating common-sense reasoning capabilities. We evaluate on the common-sense reasoning tasks specified by the official implementation, which includes 9 tasks/datasets: ARC-Challenge, ARC-Easy, BoolQ, HellaSwag, Lambada, PIQA, SocialIQA, Wikitext, and Winograd. The scores we report in Table~\ref{tab:downstream-performance} are the average accuracy score (excluding Wikitext, which is evaluated with perplexity). When available, we use the normalized accuracy scores instead of raw accuracy scores.

\paragraph{Retrieval Task}

We report the average accuracy of the synthetic S-NIAH tasks from RULER~\citep{ruler}, which tests the model's ability to retrieve a certain ``needle'' (i.e., some special information) from a large body of irrelevant text.

\subsection{Context Length Extension by Post-Training}

\begin{figure}[!t]
    \centering
    \includegraphics[width=0.98\linewidth]{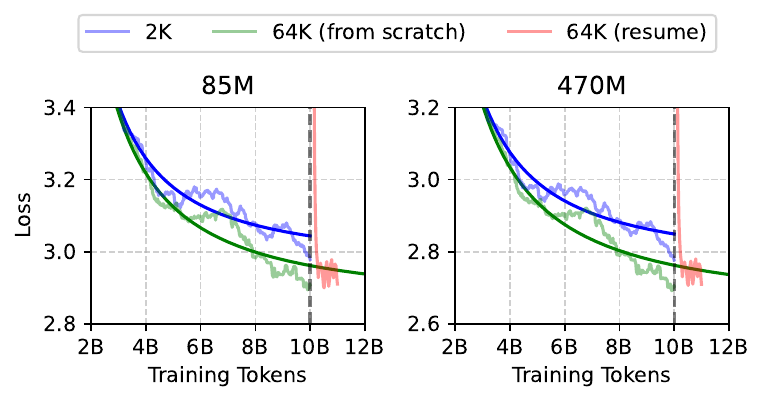}
    \caption{The loss curves of a model with 2K context length adapted to 64K through post-training compared to a model trained with 64K from scratch.}
    \label{fig:post-training}
\end{figure}

LLMs are typically trained on shorter sequences in practice, followed by adaptation to longer contexts using a smaller amount of data tailored to the target context length. To ensure the validity of our conclusions in such training scenarios, we adapted a checkpoint initially trained with a 2K context length to a 64K context length through continual pretraining. This adapted model was then compared to a model trained from scratch with a 64K context length. As illustrated in Figure~\ref{fig:post-training}, the adapted model rapidly converges toward the performance of the model trained from scratch with a 64K context length. This indicates that, with sufficient post-training, the loss of the adapted model approaches that of a model trained entirely from scratch. Consequently, our findings regarding inference costs and the relationship between loss, context length, and head configuration remain applicable to post-training scenarios.

\section{AI Assistance in Research and Writing}

We have used AI for code completion during implementation and grammar-check during paper-writing. We do not explicitly instruct AI to write any part of this paper.

\end{document}